\documentclass[letterpaper, 10 pt, conference]{ieeeconf}
\usepackage[T1]{fontenc}
\usepackage{anyfontsize}

\IEEEoverridecommandlockouts
\usepackage{xcolor}
\usepackage{amsmath}
\usepackage{algorithm}
\usepackage{algpseudocode}
\usepackage{amsfonts}
\usepackage{graphicx}
\usepackage{hyperref}
\usepackage{xspace}
\usepackage{soul}
\usepackage{subcaption}
\usepackage{wrapfig}
\usepackage{booktabs} 
\usepackage{multirow}

\newcommand{\methodName}{\textsc{$\arg$-Vu}\xspace}

\title{\LARGE \bf \methodName: \ul{A}ffordance \ul{R}easoning with Physics-Aware 3D \ul{G}eometry for \ul{V}isual \ul{U}nderstanding in Robotic Surgery}

\author{
Nan Xiao$^{1}$, Yunxin Fan$^{2}$, Farong Wang$^{1}$, and Fei Liu$^{1}$\\
$^{1}$University of Tennessee, Knoxville, TN, USA\\
$^{2}$Stanford University, Palo Alto, CA, USA\\
{\tt\small nxiao4@vols.utk.edu, yunxin6@stanford.edu, fwang31@vols.utk.edu, fliu33@utk.edu}
}
\begin{document}
\maketitle
\thispagestyle{empty}
\pagestyle{empty}

\begin{abstract}
Affordance reasoning provides a principled link between perception and action, yet remains underexplored in surgical robotics, where tissues are highly deformable, compliant, and dynamically coupled with tool motion. We present \methodName, a physics-aware affordance reasoning framework that integrates temporally consistent geometry tracking with constraint-induced mechanical modeling for surgical visual understanding. Surgical scenes are reconstructed using 3D Gaussian Splatting (3DGS) and converted into a temporally tracked surface representation. Extended Position-Based Dynamics (XPBD) embeds local deformation constraints and produces representative geometry points (RGPs) whose constraint sensitivities define anisotropic stiffness metrics capturing the local constraint-manifold geometry. Robotic tool poses in $\mathrm{SE}(3)$ are incorporated to compute rigidly induced displacements at RGPs, from which we derive two complementary measures: a physics-aware compliance energy that evaluates mechanical feasibility with respect to local deformation constraints, and a positional agreement score that captures motion alignment (as kinematic motion baseline). Experiments on surgical video datasets show that \methodName yields more stable, physically consistent, and interpretable affordance predictions than kinematic baselines. These results demonstrate that physics-aware geometric representations enable reliable affordance reasoning for deformable surgical environments and support embodied robotic interaction.
\end{abstract}

\section{Introduction}

Affordance reasoning concerns an agent’s ability to perceive and interpret actionable possibilities offered by the environment. Originally introduced in ecological psychology \cite{Chemero_2003}, affordances describe how objects or surfaces enable interactions such as grasping, cutting, or supporting, depending jointly on environmental properties and the agent’s capabilities. In robotics, affordance reasoning provides a principled bridge between perception and control: rather than treating perception as passive recognition, it frames visual observations in terms of physically meaningful opportunities for interaction \cite{Paola_2021}. Most existing approaches to affordance representation rely primarily on vision-driven modeling cues, such as contact points or grasping poses \cite{Janak_2025_TRO, Soroush_2025_ICRA, Michael_2025_ICRA, Bahl_2023_CVPR}, to guide manipulation reasoning. The central challenge, however, lies in developing representations that remain interpretable, physically grounded, and reliable for computation in complex real-world environments.

Despite extensive study in general robotics, affordance reasoning has not been systematically investigated in surgical robotics. Most recent work in surgical perception focuses on subtasks such as tool tracking \cite{Christopher_2024_ICRA, Florian_2022}, gesture recognition \cite{Amsterdam_2022, Runzhuo_2022_npj}, and scene reconstruction \cite{Shan_2024_IROS, Yan_Deform3DGS_MICCAI2024}. While these approaches enable recognition, localization, and geometric modeling, they do not explicitly address how observed tissue motion relates to actionable surgical decisions. In particular, they lack a principled linkage between visual observations and affordance-level reasoning, such as assessing tissue manipulability, evaluating whether deformation aligns with an intended surgical action, or determining whether candidate motions satisfy local mechanical constraints. Bridging this gap requires representations that move beyond recognition toward physically grounded reasoning about tool-tissue interaction.

This limitation is especially critical in surgical environments, which differ fundamentally from rigid-object manipulation. The operative field consists of highly deformable and compliant tissues, and tool actions such as grasping, retracting, or dissecting must respect underlying biomechanical properties and constraints \cite{Zixin_2025_TMI, Fei_2021_ICRA, Samuel_2026, Xiao_2024_ICRA}. Unlike rigid objects, tissues deform anisotropically, exhibit nonlinear and deformation-dependent responses, and are frequently occluded by instruments or surrounding anatomy. As a result, visual or geometric cues alone are insufficient for robotic assistance, since they cannot reliably indicate whether an observed motion is mechanically feasible or consistent with safe interaction. Addressing this challenge requires representations that support visual understanding grounded in physics-aware reasoning by capturing deformation structure, respecting anatomical constraints, and remaining computationally tractable for real-time analysis.

\begin{figure}[!t]
    \centering
    \includegraphics[width=1.0\linewidth]{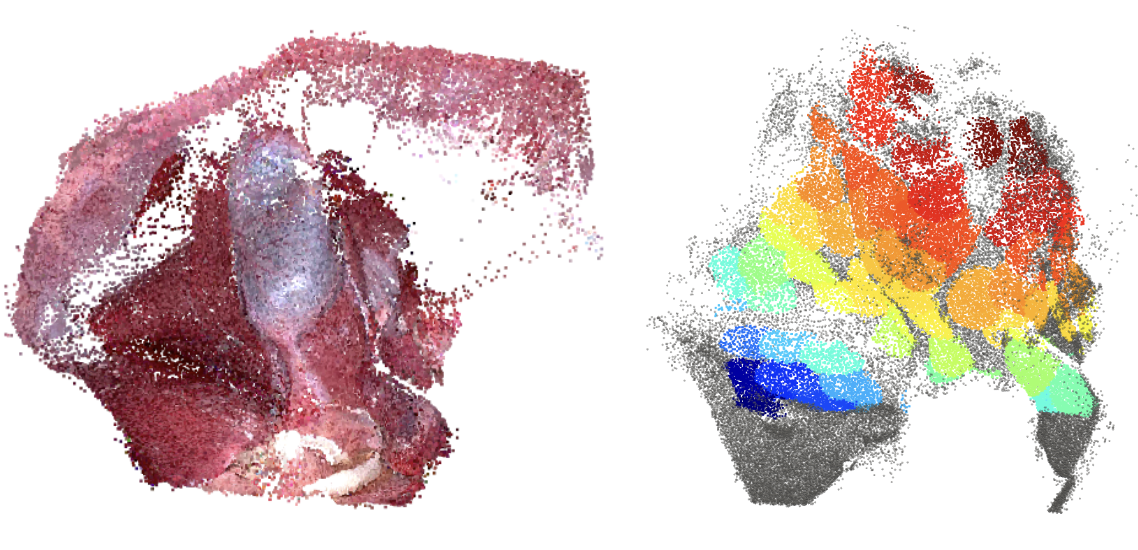}
    \caption{Left: tracked surface geometry reconstructed from endoscopic video. Right: affordance score map (red indicates higher actionability). Our framework estimates physically plausible and actionable tool-tissue interaction regions rather than only reconstructing geometry.}
    \vspace{-1.2em}
    \label{fig:cover_photo}
\end{figure}

\begin{figure*}[!htb]
    \centering
    \includegraphics[width=0.98\textwidth]{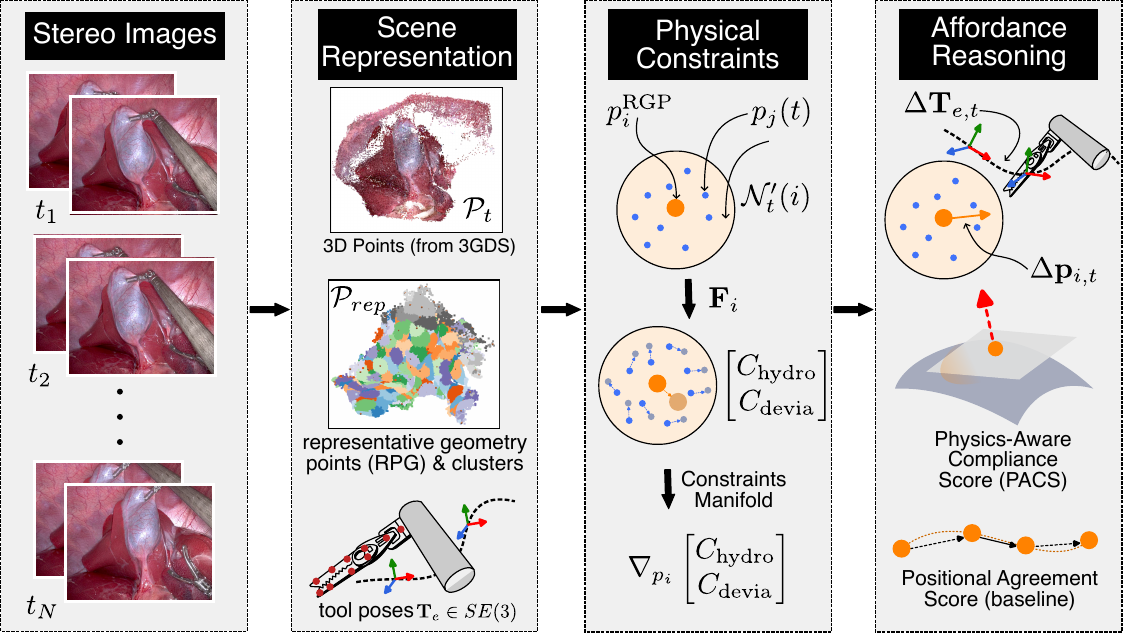}
    \caption{Overall framework of \methodName. Th 3D point cloud from 3DGS provides a temporally tracked surface representation. XPBD enforces local constraints to produce representative geometry point (RGPs) with anisotropic stiffness structure. Tool poses in $\mathrm{SE}(3)$ yield induced displacements used for PACS (physical feasibility) and PAS (motion alignment, as kinematic motion baseline) for affordance reasoning.}
    \label{fig:overall}
\end{figure*}

To address this need, we introduce a physics-aware affordance reasoning framework that integrates temporally consistent surface geometry tracking with constraint-induced mechanical modeling. Rather than performing full biomechanical simulation, our goal is to compute a lightweight and interpretable affordance measure by extracting a physically grounded deformation manifold from tracked tissue motion under explicit deformation constraints. This formulation assigns clear physical meaning to affordance by quantifying whether observed or candidate tool-tissue motions are mechanically feasible under local deformation constraints, thereby enabling physically grounded visual understanding of interaction.

\subsection{Contributions}
Building on our prior work on first-order position-based methods using deformation gradients \cite{Samuel_2025_ISMR, XPBD} and stereo-view geometry reconstruction via 3DGS \cite{Kerbl2023GaussianSplatting}, we propose a unified framework for surgical affordance reasoning that jointly captures (i) temporally consistent surface motion from video and (ii) anisotropic local compliance derived from constraint sensitivity. Our key contributions are as follows:

\begin{itemize}
    \item A geometry-centric representation that integrates stereo-view tracking and incorporates physics-aware compliance constraints for consistent modeling of the tissue surface.
    \item A constraint manifold grounded stiffness formulation with Jacobian-based sensitivity analysis for physically understanding about surgical tissue deformation.
    \item A physics-aware affordance formulation that maps end-effector motion in $\mathrm{SE}(3)$ to local tissue response, yielding spatially interpretable maps of feasible tool-tissue interactions.
\end{itemize}

\section{Related Works}
\label{sec:rw}

\subsection{Visual and Physical Modeling of Surgical Environments}
Neural scene representations have recently advanced the ability to model anatomy and appearance in surgical environments. Methods such as EndoNeRF \cite{wang2022neural}, EndoGaussian \cite{Yifan_2025_TMI}, and deformable 3DGS variants \cite{Yan_Deform3DGS_MICCAI2024} extend implicit or Gaussian-based representations to capture non-rigid motions of soft tissue in endoscopic videos. These approaches demonstrate strong visual fidelity and enable geometry reconstruction under deformation, yet they remain largely appearance-driven and do not explicitly incorporate physical constraints governing tissue mechanics (e.g., compliance and anisotropic deformation under tool-induced motion).

In parallel, soft tissue simulation has traditionally relied on FEM and surgical simulation frameworks such as SOFA \cite{Faure_2012_SOFA} for high-fidelity deformation. While accurate, FEM-based approaches typically require volumetric meshing and material parameters, and can be computationally expensive for real-time perception-driven reasoning. Particle-based approaches such as PBD \cite{PBD} provide greater efficiency and robustness, but their stiffness is coupled to solver iterations. XPBD \cite{XPBD} decouples stiffness from iteration count via compliance parameters, retaining the real-time advantages of PBD while enabling more physically grounded modeling. Unlike simulation-only approaches, our work embeds constraint-based physical sensitivity into a 3DGS-derived tracked surface 3D representation to support actionable affordance reasoning.

\subsection{Affordance Reasoning in Robotics}
Affordance reasoning links perception to action \cite{Paola_2021}. Many works learn affordances for grasping and manipulation, often assuming rigid objects \cite{Janak_2025_TRO, Soroush_2025_ICRA, Michael_2025_ICRA}. Recent research extends affordance reasoning to deformable objects such as cloth \cite{Wanjun_2025, WuRuihai_2025_CVPR}, but such approaches are typically driven by visual cues and may lack explicit physical grounding. Surgical tissues amplify these challenges due to compliance, heterogeneity, and dynamic coupling with tool actions. Our method departs from prior work by embedding physics-aware compliance constraints into affordance estimation, enabling physically grounded and interpretable robot end-effector actions for surgical environments.

\section{Methodology}
\label{sec:method}



\subsection{Scene Representation via Geometry Tracking}
\label{sec:3DGS}

We represent the surgical scene using a temporally consistent geometric reconstruction derived from endoscopic video. Specifically, we leverage 3D Gaussian Splatting (3DGS) \cite{Kerbl2023GaussianSplatting} as a backend for dense stereo-view reconstruction and tracking, using it primarily to obtain stable geometric observations rather than as a modeling primitive. Following \cite{Hay_Online_MICCAI2024}, we convert the optimized reconstruction into a temporally tracked surface representation $\mathcal{P}_t = \{p_i(t)\}_{i=1}^N$, 
by associating each tracked point with the corresponding reconstructed center across frames.

The key advantage of this geometry-centric formulation is that it provides (i) a dense and view-consistent surface representation directly from endoscopic video and (ii) improved temporal stability under specularities, partial occlusions, and viewpoint changes. This stability is essential because our physics-aware formulation (Section~\ref{sec:physics_embedding}) operates on local neighborhoods and is sensitive to high-frequency tracking noise. By enforcing stereo-view consistency through differentiable rendering, the reconstruction backend substantially reduces temporal jitter in $\mathcal{P}_t$ compared with naive frame-wise point tracking.

Importantly, subsequent physical reasoning depends only on the tracked geometry $\mathcal{P}_t$ rather than the underlying radiance representation. The tracked points serve as representative material locations from which local neighborhoods are constructed and constraint-based mechanical analysis is performed.

\subsection{Physics-Aware Constraints Embedding}
\label{sec:physics_embedding}
To move beyond appearance-only modeling, we embed physics-aware constraints into the tracked surface geometry reconstructed from endoscopic video. Rather than relying on the radiance representation itself, we map each representative location to a tracked point augmented with compliance (i.e., stiffness) parameters derived from XPBD \cite{XPBD}. This formulation decouples stiffness characterization from solver iterations and enables anisotropic deformation modeling grounded in local constraint structure.

Our objective is not full biomechanical simulation such as in \cite{Fei_2021_ICRA, Samuel_2026}; instead, we seek to extract a lightweight yet physically meaningful notion of \emph{local compliance} directly from tracked surface motion. This yields an interpretable physical representation that supports downstream affordance reasoning: regions that appear to move with the tool but violate local mechanical feasibility should not be considered highly actionable.

\subsubsection{Representative Geometry Points (RGPs)}

To reduce computation, we operate on a sparse subset of \textbf{representative geometry points (RGPs)} $\mathcal{P}_{rep} = \{p_i^{\text{RGP}}\}_{i=1}^{N_{rep}} \subset \mathcal{P}_0$ from the first frame (frame 0) that serve as fixed control locations for local mechanical modeling. The RGP set is selected using radius-based clustering with radius $r_{cluster}$ and a minimum neighborhood size $N_{min}$. The identities of the RGPs remain fixed over time.

Given the fixed RGPs $\mathcal{P}_{rep}$, we construct a time-varying \textbf{neighborhood map} $\mathcal{N}_t$ to account for motion and local rearrangements. For each $p_i^{\text{RGP}} \in \mathcal{P}_{rep}$, we define
\begin{equation}
    \mathcal{N}_t(i) = \left\{\, p_j(t) \in \mathcal{P}_t \;\middle|\; \| p_j(t) - p_i^{\text{RGP}}(t) \| \le r_{cluster} \right\}.
\end{equation}
This dynamic neighborhood update captures local spatial relationships at each time step without requiring mesh construction and remains robust to large motion and topology changes common in surgical scenes.

Importantly, radius-based clustering induces \emph{soft boundaries}: RGPs act as sparse control locations for estimating local mechanical quantities and do not impose a hard partition of tissue. Neighborhoods may overlap, and each tracked point in $\mathcal{P}_t$ can influence multiple RGPs through continuous kernel weights $w_{ij}$ (Eq.~\ref{eq:deformation_gradient}), yielding smooth spatial variation in local estimates. Adaptive clamping in the solver further suppresses discontinuous corrections caused by tracking noise.

\subsubsection{Segmentation-Aware Neighborhood Filtering}

A primary challenge in mesh-free neighborhood methods is defining meaningful neighborhoods when multiple anatomical structures are in contact. While $\mathcal{N}_t(i)$ captures local spatial relationships, it is agnostic to anatomical boundaries. We introduce \textbf{dynamic filtering} using per-frame 2D segmentations. Each 3D point is projected into the image plane and assigned a label $S(p,t)$. The filtered neighborhood is defined as
\begin{equation}
\label{eq:seg_filter}
    \mathcal{N}'_t(i) = \left\{\, p_j(t) \in \mathcal{N}_t(i) \;\middle|\; S(p_j,t) = S\!\left(p_i^{\text{RGP}},t\right) \right\}.
\end{equation}

This prevents physically implausible constraints across distinct anatomical structures and supports heterogeneous mechanical behavior.

\begin{figure}[t]
    \centering
    \includegraphics[width=0.48\textwidth]{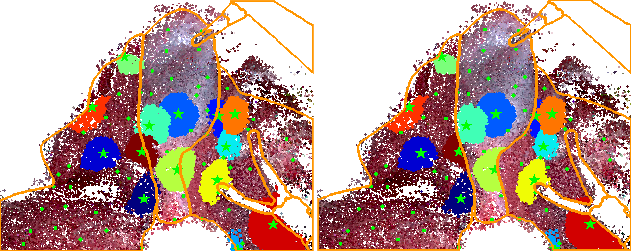}
    \caption{Neighborhood definitions. \textbf{Left}: radius neighborhoods may cross anatomical boundaries. \textbf{Right}: segmentation-aware filtering constrains neighborhoods within anatomical regions, improving physical plausibility near organ boundaries.}
    \label{fig:seg_neighbor}
\end{figure}

Fig.~\ref{fig:seg_neighbor} illustrates a qualitative comparison: the naive radius neighborhood (left) may cross anatomical boundaries, whereas segmentation-aware filtering (right) restricts neighborhoods to semantically consistent regions. This distinction is especially important in surgical settings, where nearby tissues may exhibit fundamentally different compliance responses, and mixing them can lead to inaccurate local estimates and unstable constraints.

\subsubsection{XPBD-Based Deformation Constraint Modeling}

We employ a quasi-static XPBD solver \cite{Samuel_2025_ISMR} to model tissue elastic response by solving for equilibrium at each step. To obtain a stable reference configuration, we construct a smoothed \textbf{rest state} for each tracked point (including RGPs), using temporal averaging over a window $\mathcal{T}_i = [t_{\text{appr}}(i),\, t_{\text{appr}}(i)+\tau]$:
\begin{equation}
    \bar{p}_j^0 = \frac{1}{|\mathcal{T}_i|} \sum_{t' \in \mathcal{T}_i} p_j(t').
\end{equation}
This smoothing reduces high-frequency tracking noise and yields a more reliable reference neighborhood for deformation estimation.

For each RGP $p_i^{\text{RGP}}$, we estimate the local deformation gradient $\mathbf{F}_i$ from its filtered neighborhood $\mathcal{N}'_t(i)$ via weighted least squares:
\begin{equation}
\label{eq:deformation_gradient}
\begin{split}
    \mathbf{F}_i  & = \underset{\mathbf{F}}{\arg\min} \sum_{p_j(t) \in \mathcal{N}'_t(i)}   \\
    & w_{ij} \left\| \big(p_j(t) - p_i^{\text{RGP}}(t)\big) - \mathbf{F}\big(\bar{p}_j^0 - \bar{p}_i^0\big) \right\|^2,
\end{split}
\end{equation}
where $w_{ij}=\exp(-\|\bar{p}_j^0-\bar{p}_i^0\|^2/\sigma^2)$ assigns greater weight to closer neighbors. Intuitively, $\mathbf{F}_i$ summarizes local tissue behavior as the best linear mapping from rest-space offsets to current offsets. 

We enforce elasticity through two constraints derived from $\mathbf{F}_i$:
\begin{itemize}
    \item \textbf{Hydrostatic constraint:} $C_{\text{hydro}} = \det(\mathbf{F}_i) - 1$, encouraging near-volume preservation.
    \item \textbf{Deviatoric constraint:} $C_{\text{devia}} = \mathrm{tr}(\mathbf{F}_i^\top \mathbf{F}_i) - 3$, penalizing shear and excessive distortion.
\end{itemize}

We solve the resulting constraint system using Gauss–Seidel updates with adaptive clamping (98th percentile) to suppress extreme corrections caused by outlier tracking errors, improving robustness under specularities and occlusion.

\subsubsection{Constraint-Induced Stiffness Metric at Representative Geometry Points (RGPs)}

\begin{figure}[!htb]
\centering
\includegraphics[width=0.99\linewidth]{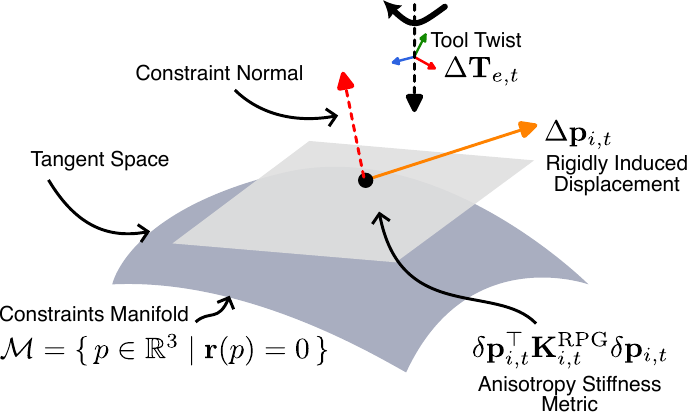}
\caption{
Geometric interpretation of the constraint-induced stiffness metric.
Local deformation constraints define a manifold $\mathcal{M}$ whose tangent space represents admissible deformation directions.
The stiffness metric $\mathbf{K}_i^{\mathrm{RGP}}$ induces an ellipsoid that visualizes anisotropic resistance to motion.
Long axes correspond to compliant tangent directions, whereas short axes correspond to stiff constraint-normal directions.
A tool-induced displacement $\Delta p_i$ at the representative point $p_i$ is evaluated through this metric.
}
\label{fig:constraint_manifold}
\end{figure}

At each representative geometry point (RGP) $p_i$, the local deformation behavior is governed by physically motivated constraints that implicitly define a constraint manifold
\[
\mathcal{M} = \{\, p \in \mathbb{R}^3 \mid \mathbf{r}(p)=0 \,\},
\]
where the residual vector is given by
$\mathbf{r}_i = [C_{\mathrm{hydro}},\, C_{\mathrm{devia}}]^\top$.
Points on $\mathcal{M}$ satisfy the local deformation constraints, while admissible deformation motions correspond to directions tangent to this manifold.

Let
$\mathbf{J}_i^{\mathrm{RGP}} = \partial \mathbf{r}_i / \partial p_i \in \mathbb{R}^{2\times3}$
denote the constraint sensitivity Jacobian.
A displacement $\delta p$ preserves the constraints to first order when
$\mathbf{J}_i^{\mathrm{RGP}} \delta p = 0$,
which defines the tangent space
\[
T_{p_i}\mathcal{M} = \mathrm{Null}(\mathbf{J}_i^{\mathrm{RGP}}).
\]
These tangent directions correspond to locally admissible deformation modes.
In contrast, motions in the range of $\mathbf{J}_i^{\mathrm{RGP}\top}$ alter the constraint values and therefore represent constraint-normal directions that are mechanically resisted.

To quantify this anisotropic resistance, we introduce a constraint-induced stiffness metric
\begin{equation}
\label{eq:xpbd_stiffness_metric}
\mathbf{K}_i^{\mathrm{RGP}}
=
\mathbf{J}_i^{\mathrm{RGP}\top}
\mathbf{J}_i^{\mathrm{RGP}}
+
\varepsilon \mathbf{I}_3 ,
\end{equation}
where $\varepsilon>0$ ensures numerical stability.
The quadratic form $\delta p^\top \mathbf{K}_i^{\mathrm{RGP}} \delta p$ measures the first-order deviation of a displacement $\delta p$ from the constraint manifold.
Displacements aligned with constraint-normal directions yield large energy and correspond to stiff deformation modes, whereas motions within the tangent space yield low energy and correspond to compliant directions.

The anisotropy of this metric can be visualized through the ellipsoid defined by
$\delta p^\top \mathbf{K}_i^{\mathrm{RGP}} \delta p = c$,
which represents a constant-energy contour.
Short ellipsoid axes correspond to stiff constraint-normal directions, while long axes align with compliant tangent directions of the constraint manifold.
Fig.~\ref{fig:constraint_manifold} illustrates this geometric interpretation.

\subsection{Physics-Aware Compliance Energy (PACE) as Interaction Affordance}

We formulate interaction affordance directly as a physics-aware compliance energy that quantifies whether an induced rigid end-effector motion is mechanically compatible with local tissue deformation constraints. The core idea is to (i) map a tool motion in $\mathrm{SE}(3)$ to the instantaneous displacement it would induce at a tissue-attached representative geometry point (RGP), and (ii) evaluate that displacement under a constraint-induced compliance model that encodes which local deformation directions are mechanically compliant (“easy”) and which are stiff (“hard”).

The resulting scalar energy provides a physically grounded measure of interaction compatibility: low energy indicates motion aligned with compliant directions and thus mechanically supported by the tissue, whereas high energy indicates conflict with local constraint structure. This energy can be visualized spatially across RGPs and tracked temporally during execution to assess interaction stability.

\subsubsection{Rigidly Induced Displacement}

Let the end-effector pose at step $t$ be $\mathbf{T}_{e,t} = [\mathbf{R}_t \ \mathbf{o}_t] \in \mathrm{SE}(3)$, with origin $\mathbf{o}_t$.
Define the discrete increments
$\Delta \mathbf{R}_t = \mathbf{R}_{t+1}\mathbf{R}_t^\top$ and
$\Delta \mathbf{o}_t = \mathbf{o}_{t+1} - \mathbf{o}_t$.
Let the incremental twist (i.e., finite pose increments) be
$\boldsymbol{\xi}_t = [\mathbf{v}_t^\top,\boldsymbol{\omega}_t^\top]^\top$,
where $\mathbf{v}_t = \Delta \mathbf{o}_t$ and
$\boldsymbol{\omega}_t = \mathrm{Log}(\Delta \mathbf{R}_t)^\vee$.

Then the rigidly induced displacement (i.e., finite displacement over the time step) at RGP $\mathbf{p}_i(t)$ is
\begin{equation}
\Delta \mathbf{p}_{i,t}
=
\mathbf{v}_t
+
\boldsymbol{\omega}_t \times \big(\mathbf{p}_i(t) - \mathbf{o}_t\big).
\label{eq:tool_induced_displacement}
\end{equation}

Equivalently, defining the point Jacobian
$\mathbf{J}_{\mathrm{pt}}(\mathbf{p}_i(t),\mathbf{o}_t)
=
\begin{bmatrix}
\mathbf{I}_3 & -[\mathbf{p}_i(t)-\mathbf{o}_t]_\times
\end{bmatrix}$,
we have
\begin{equation}
\Delta \mathbf{p}_{i,t}
=
\mathbf{J}_{\mathrm{pt}}(\mathbf{p}_i(t),\mathbf{o}_t)\boldsymbol{\xi}_t.
\end{equation}

\subsubsection{Physics-Aware Compliance Score (PACS)}

Each RGP is associated with a constraint-induced stiffness matrix $\mathbf{K}^{\mathrm{RGP}}_{i,t}$ defines the local stiffness metric of the deformation manifold.
Given the rigidly induced displacement
$\Delta \mathbf{p}_{i,t}$,
we define the \textbf{Physics-Aware Compliance Energy (PACE)}:
\begin{equation}
E_{i,t}
=
\frac{1}{2}
\Delta \mathbf{p}_{i,t}^\top
\mathbf{K}^{\mathrm{RGP}}_{i,t}
\Delta \mathbf{p}_{i,t}.
\end{equation}

This quadratic form measures how strongly the induced displacement projects onto locally stiff directions.
Small $E_{i,t}$ indicates motion aligned with compliant directions and thus mechanically admissible deformation,
whereas large $E_{i,t}$ indicates incompatibility with the local constraint structure.

We therefore define the \textbf{Physics-Aware Compliance Score (PACS)}
\begin{equation}
\mathcal{A}_{i,t}
=
- E_{i,t},
\end{equation}
so that higher values correspond to greater mechanical compatibility for the interaction.

\subsubsection{Noise-Aware and Temporal Affordance}

To mitigate pose jitter and transient contact fluctuations,
we smooth the PACE using an exponential moving average
\begin{equation}
\bar{E}_{i,t}
=
\lambda \bar{E}_{i,t-1}
+
(1-\lambda)E_{i,t},
\qquad
\lambda\in(0,1),
\end{equation}
and define the smoothed PACS
\begin{equation}
\bar{\mathcal{A}}_{i,t}
=
- \bar{E}_{i,t}.
\end{equation}
This $\bar{\mathcal{A}}_{i,t}$ would then be our affordance measure.

\section{Experiments and Results}
\label{sec:er}

\subsection{Data Processing}
We evaluate on the \textbf{StereoMIS} dataset \cite{hayoz2023pose}, focusing on a gallbladder retraction sequence:

\paragraph{Dataset details}
We evaluate our framework on a single gallbladder retraction sequence from StereoMIS \cite{hayoz2023pose}, consisting of 200 frames at 30 fps and resolution $640\times512$, of which 125 frames involve active tool-tissue interaction. Depth is computed following the online endoscopic tracking method of Hayoz \textit{et al.} \cite{Hay_Online_MICCAI2024}, where stereo disparity is estimated from RGB pairs and back-projected to metric 3D using calibrated intrinsics and baseline. The resulting depth and Gaussian-based tracking provide temporally consistent 3D tissue points and tool poses expressed in the calibrated stereo camera frame.

\paragraph{Tool pose extractions}
For each frame, we annotate $K$ keypoints on the surgical tool using LocoTrack \cite{Seokju_2024}. We lift keypoints into 3D using depth and estimate a 6-DoF tool pose $\mathbf{T}_t\in\mathrm{SE}(3)$ via PCA: the centroid is translation and the principal axis is orientation (with sign disambiguated by temporal continuity).

\subsection{XPBD-Based Physical Stiffness Validation}

To validate that the XPBD-derived stiffness metric captures
physically meaningful deformation structure, we evaluate whether
the locally compliant direction inferred from constraint sensitivity
correlates with subsequent observed tissue motion.

Recall that each representative geometry point (RGP) is associated
with an anisotropic stiffness metric $\mathbf{K}_{i,t}^{\mathrm{RGP}}$
derived from constraint sensitivity. The eigendecomposition of
$\mathbf{K}_{i,t}^{\mathrm{RGP}}$ characterizes local deformation
anisotropy: directions corresponding to small eigenvalues represent
compliant (tangent) directions of the constraint manifold, whereas
large eigenvalues correspond to stiff constraint-normal directions.
We therefore interpret the eigenvector associated with the smallest
eigenvalue as the most compliant local deformation direction.

If the estimated stiffness structure is physically meaningful,
observed tissue motion should preferentially align with this compliant
direction in subsequent frames. To test this hypothesis, we compute
the cosine similarity between the predicted compliant direction and
the next-step observed motion direction:
\begin{equation}
\cos(\mathbf{e}_{\min}^{t}, \mathbf{u}^{t}),
\end{equation}
where $\mathbf{e}_{\min}^{t}$ denotes the eigenvector associated
with the smallest eigenvalue of $\mathbf{K}_{i,t}^{\mathrm{RGP}}$ and
\begin{equation}
\mathbf{u}^{t}
=
\frac{\mathbf{v}^{t}}{\|\mathbf{v}^{t}\|+\delta}
\end{equation}
is the normalized observed displacement direction at time $t$.
Frames with $\|\mathbf{v}^{t}\|<\epsilon$ are excluded to avoid
degenerate directions.

A high cosine similarity indicates that the direction predicted to be
compliant by XPBD coincides with the actual direction of tissue motion.
This directly tests whether the stiffness ellipsoid induced by
$\mathbf{K}_{i,t}^{\mathrm{RGP}}$ captures physically meaningful
anisotropic deformation behavior rather than incidental geometric
variation.

\begin{figure}[t]
    \centering
    \includegraphics[width=0.48\textwidth]{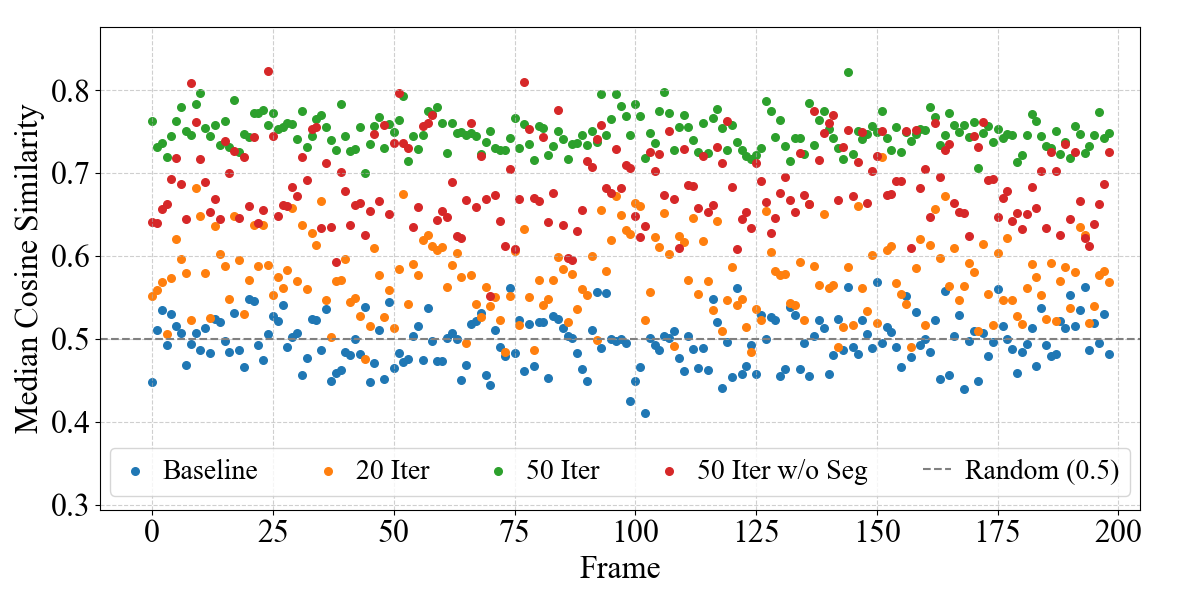}
    \caption{Median cosine similarity between observed motion direction and the principal direction of predicted uncertainty (largest eigenvector). Our XPBD-derived covariance aligns better with future motion than the velocity-persistence baseline.}
    \label{fig:curve01}
\end{figure}

We analyze performance across varying solver iteration counts to evaluate convergence behavior. Increasing XPBD iterations improves constraint satisfaction and refines the deformation gradient estimate, which in turn stabilizes the covariance estimation. As shown in Fig.~\ref{fig:curve01}, the median cosine similarity increases with solver iterations and consistently exceeds that of the baseline predictor.

\textit{Velocity-persistence baseline:}  
As a comparison, we evaluate a simple kinematic baseline that predicts the next-step motion direction as the previous observed velocity direction:
\begin{equation}
\hat{\mathbf{u}}^{t}_{\text{base}} = \frac{\mathbf{v}^{t-1}}{\|\mathbf{v}^{t-1}\|+\delta}.
\end{equation}
We compute:
\begin{equation}
\cos(\hat{\mathbf{u}}^{t}_{\text{base}},\mathbf{u}^{t}),
\end{equation}
again excluding low-magnitude frames.

Overall, this experiment demonstrates that the RGP covariance is not merely a byproduct of local geometry but encodes physically meaningful compliance directions that anticipate future tissue motion. This validates its use as a physically grounded prior in the subsequent PACS affordance formulation.

\begin{figure*}[!t]
\centering

\setlength{\tabcolsep}{1pt}
\renewcommand{\arraystretch}{1.0}

\begin{tabular}{cccccccccc}
\includegraphics[width=0.095\linewidth]{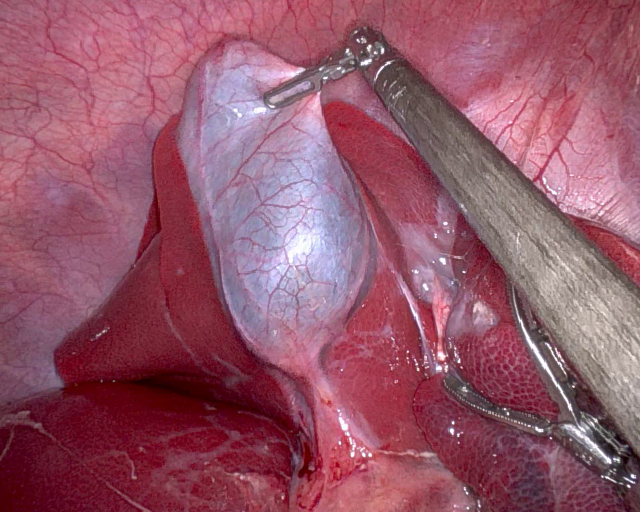} &
\includegraphics[width=0.095\linewidth]{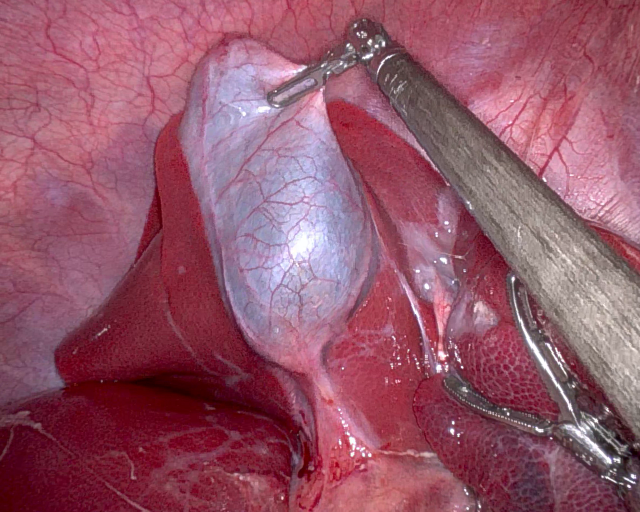} &
\includegraphics[width=0.095\linewidth]{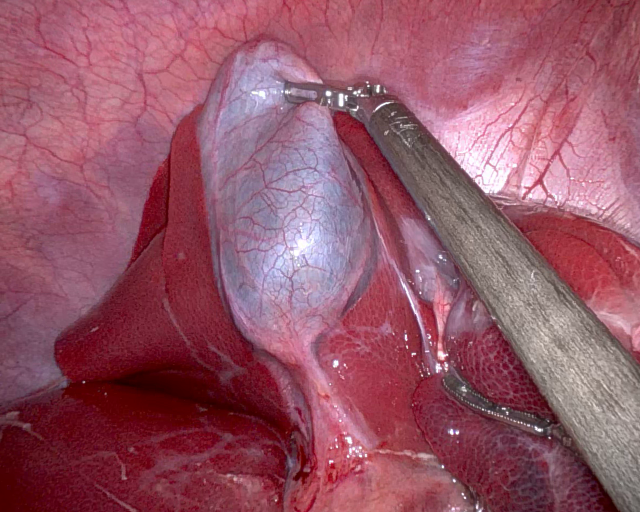} &
\includegraphics[width=0.095\linewidth]{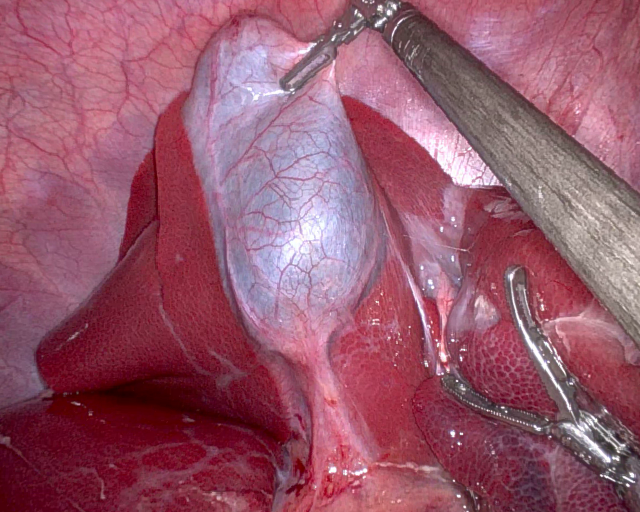} &
\includegraphics[width=0.095\linewidth]{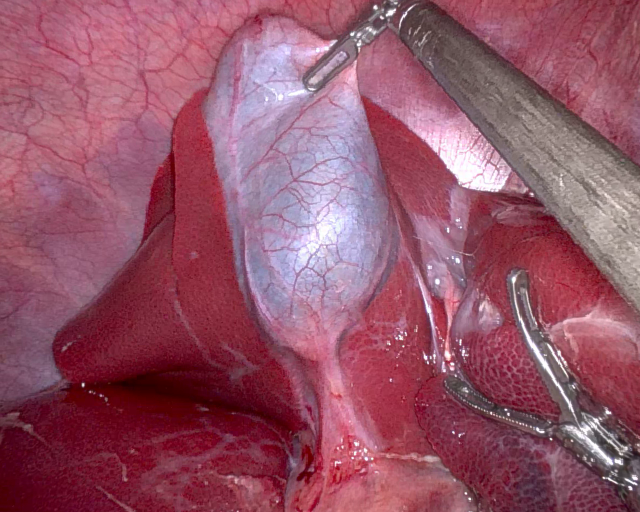} &
\includegraphics[width=0.095\linewidth]{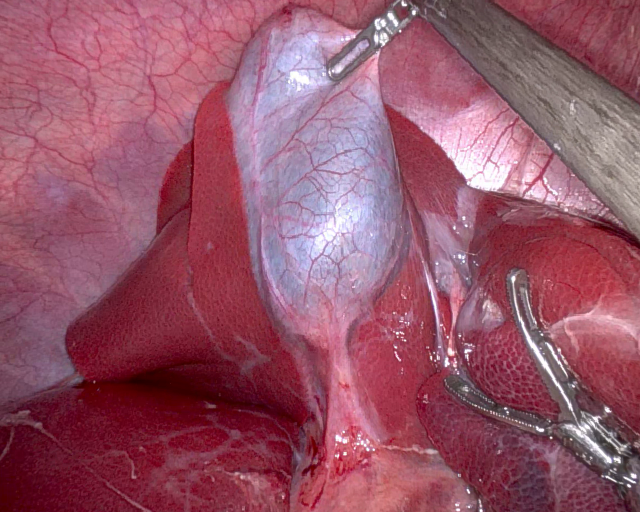} &
\includegraphics[width=0.095\linewidth]{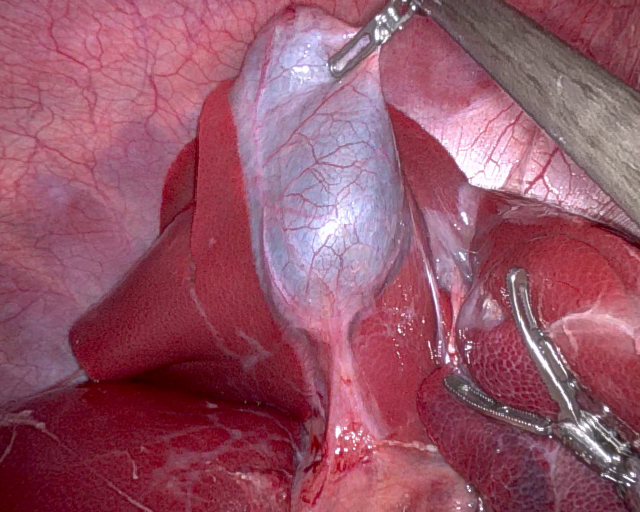} &
\includegraphics[width=0.095\linewidth]{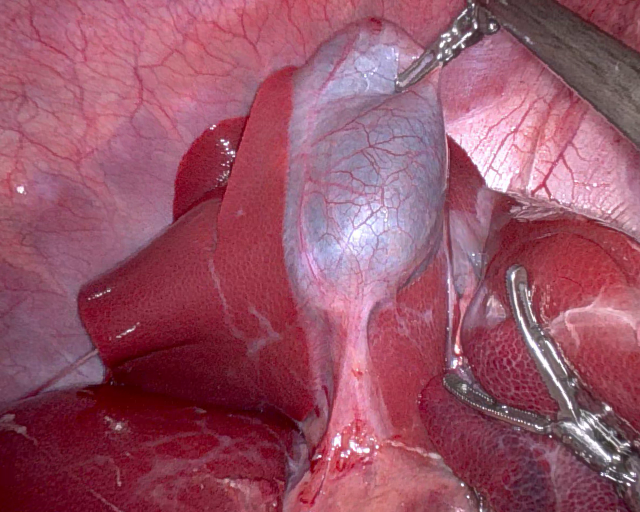} &
\includegraphics[width=0.095\linewidth]{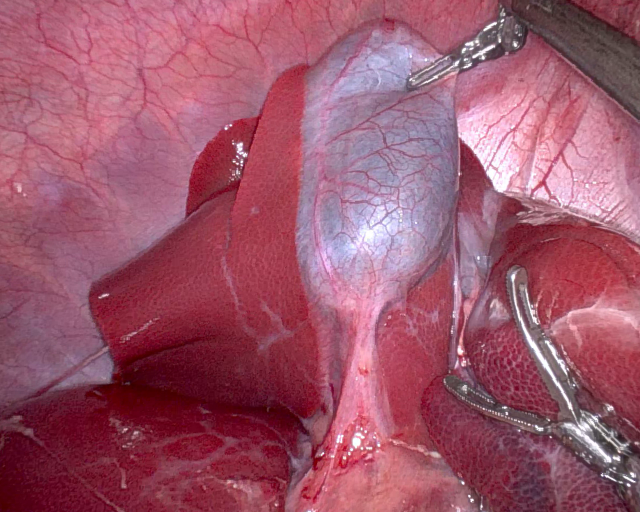} &
\includegraphics[width=0.095\linewidth]{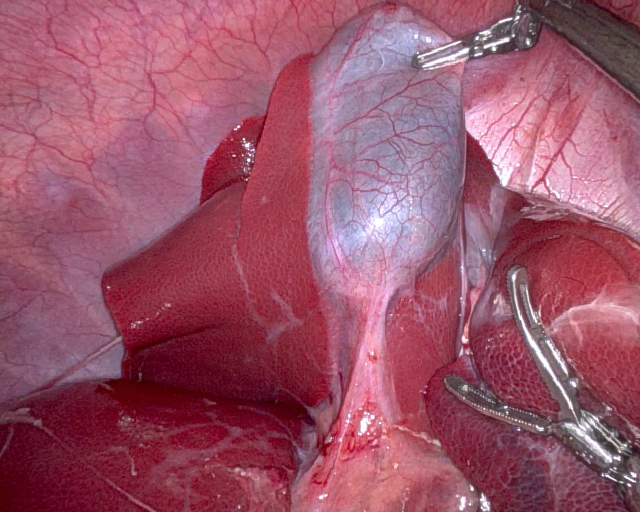}
\\[1pt]
\includegraphics[width=0.095\linewidth]{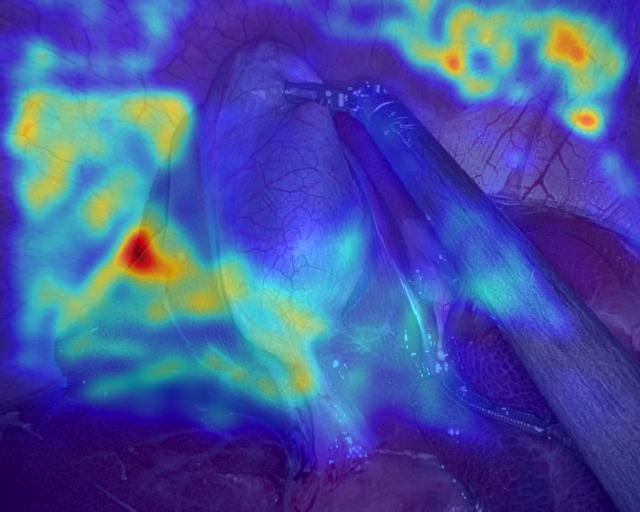} &
\includegraphics[width=0.095\linewidth]{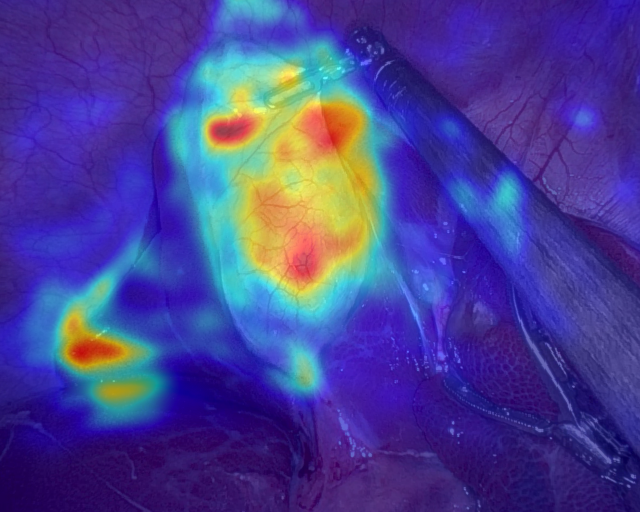} &
\includegraphics[width=0.095\linewidth]{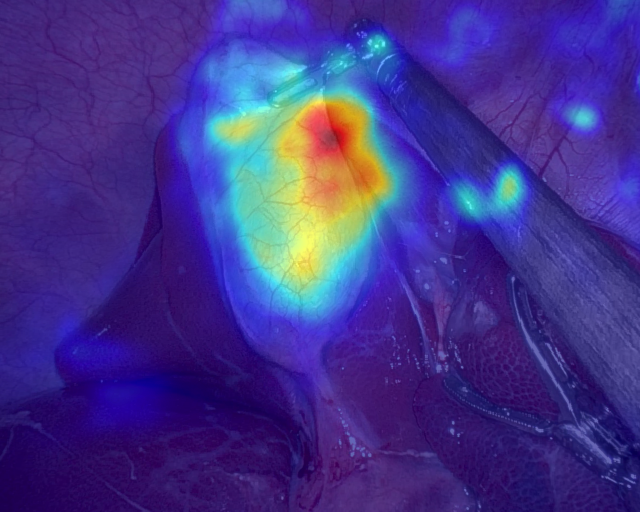} &
\includegraphics[width=0.095\linewidth]{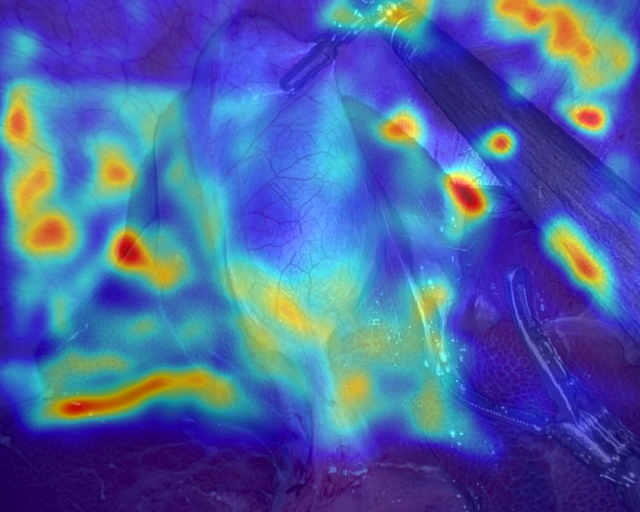} &
\includegraphics[width=0.095\linewidth]{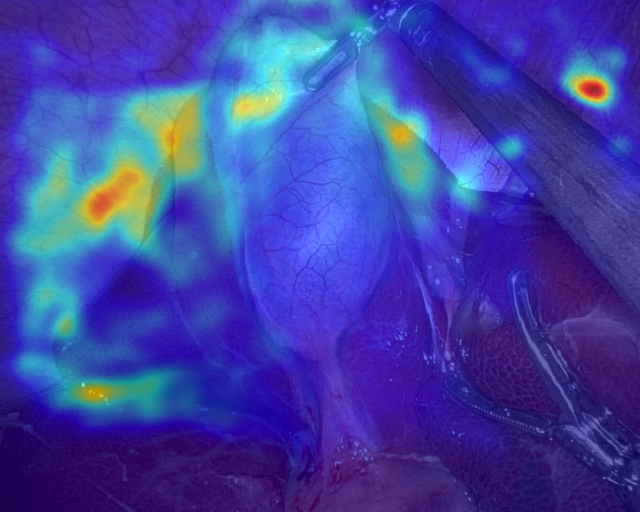} &
\includegraphics[width=0.095\linewidth]{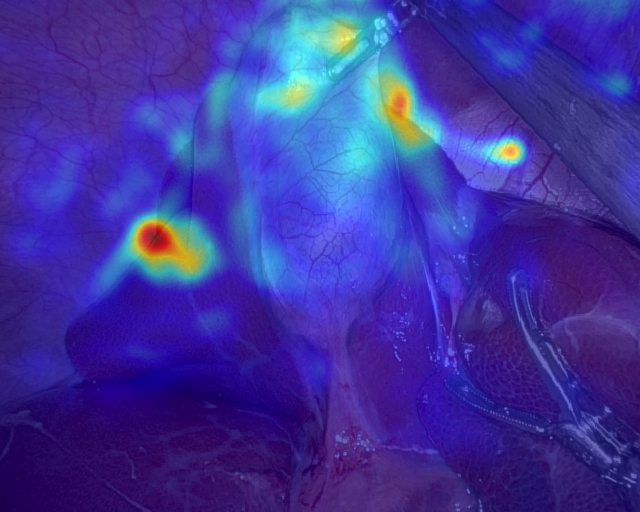} &
\includegraphics[width=0.095\linewidth]{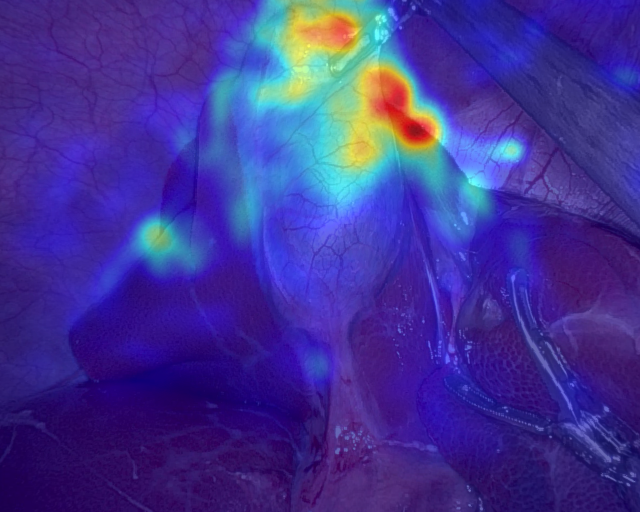} &
\includegraphics[width=0.095\linewidth]{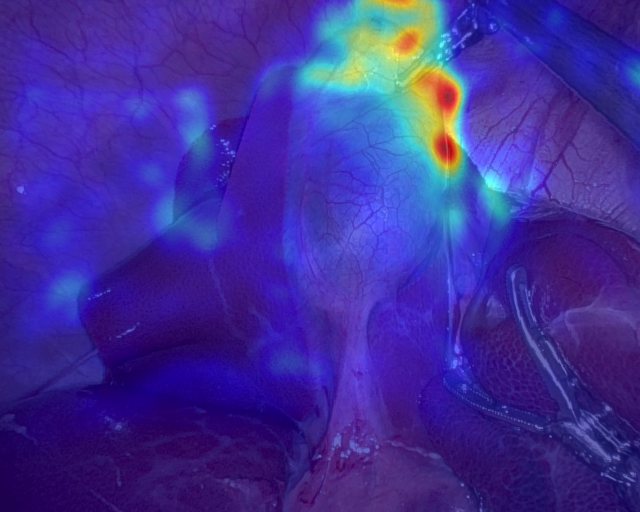} &
\includegraphics[width=0.095\linewidth]{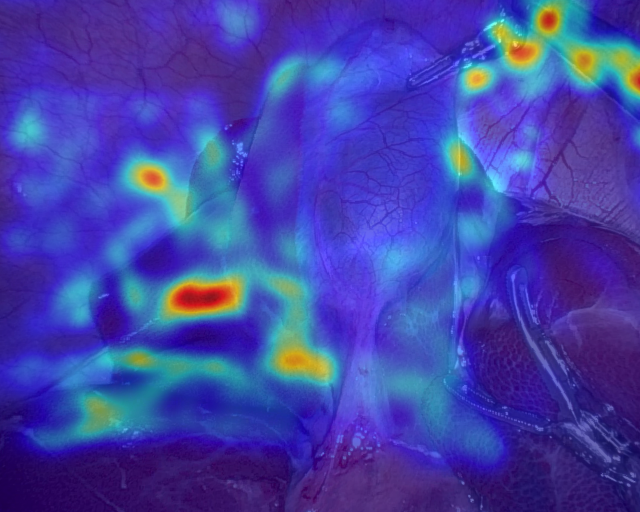} &
\includegraphics[width=0.095\linewidth]{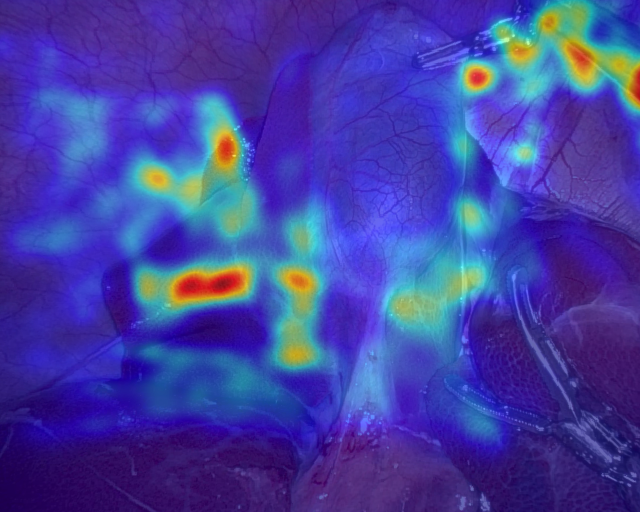}
\\[1pt]
\includegraphics[width=0.095\linewidth]{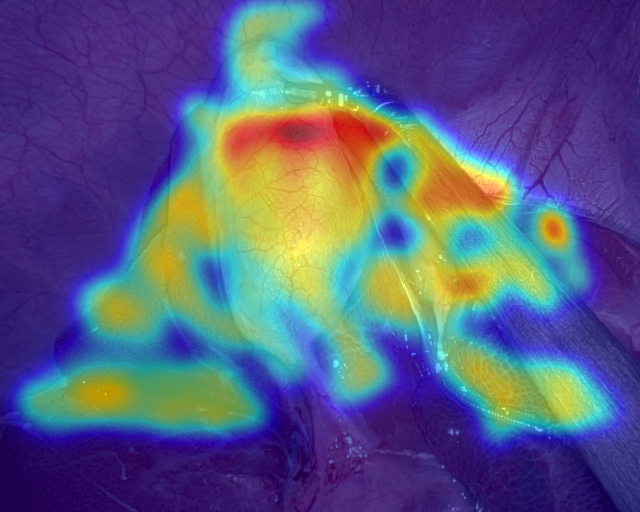} &
\includegraphics[width=0.095\linewidth]{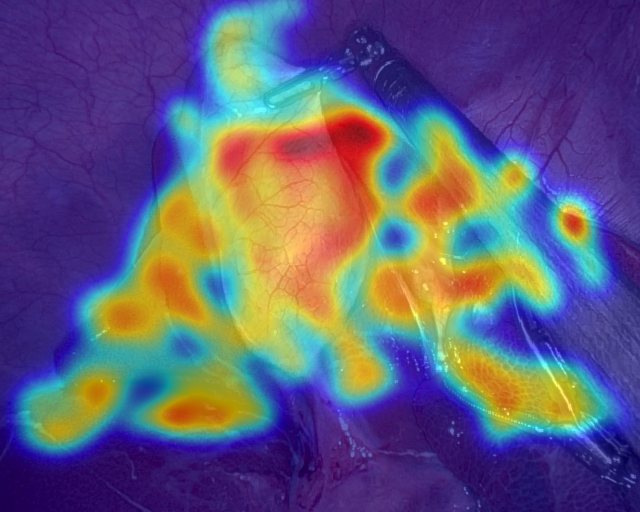} &
\includegraphics[width=0.095\linewidth]{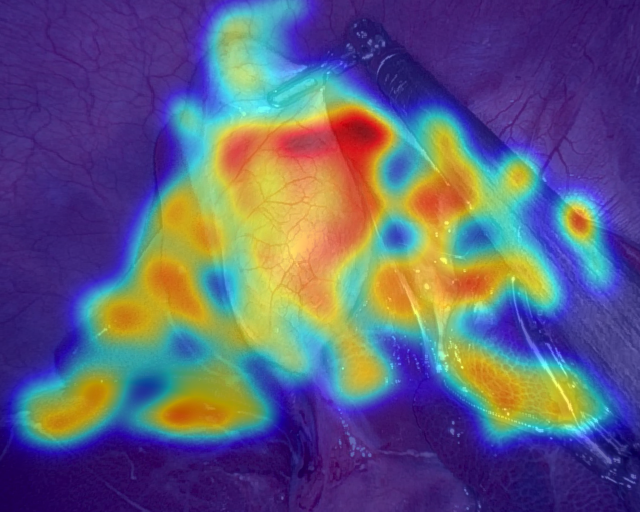} &
\includegraphics[width=0.095\linewidth]{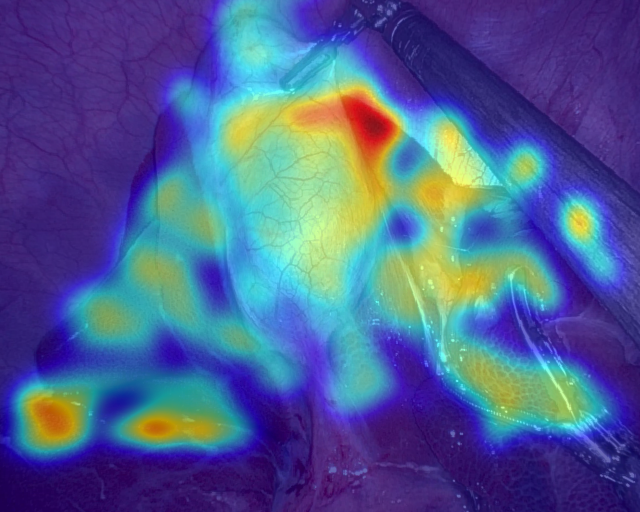} &
\includegraphics[width=0.095\linewidth]{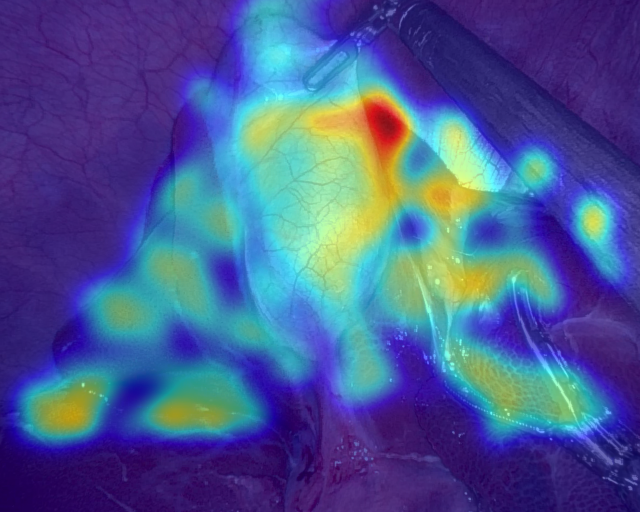} &
\includegraphics[width=0.095\linewidth]{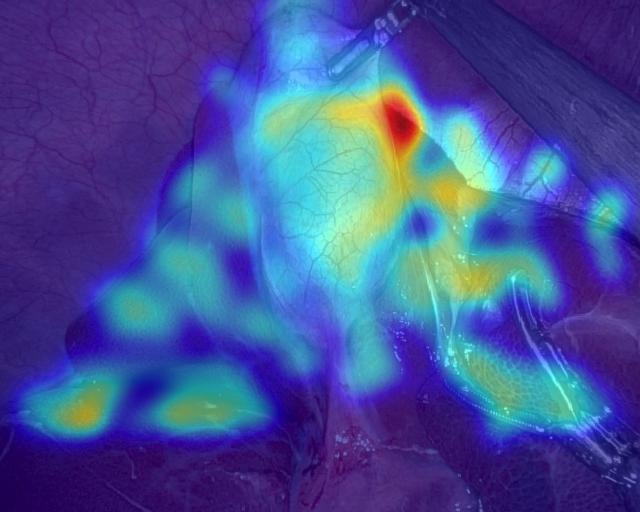} &
\includegraphics[width=0.095\linewidth]{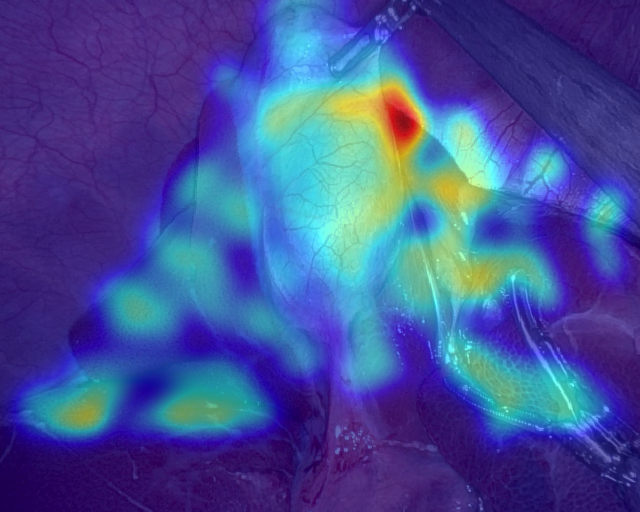} &
\includegraphics[width=0.095\linewidth]{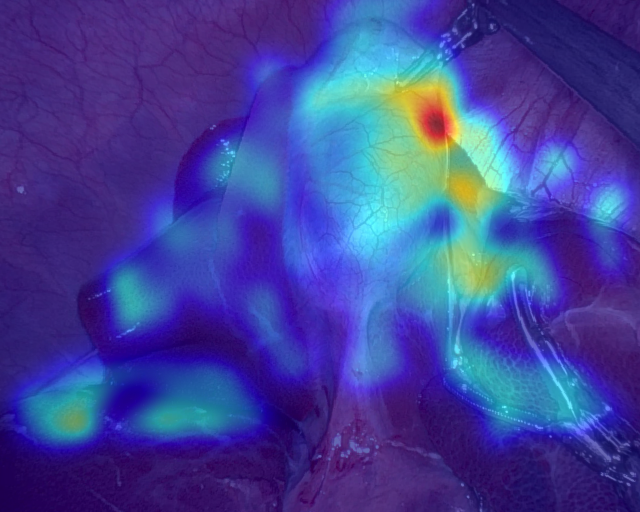} &
\includegraphics[width=0.095\linewidth]{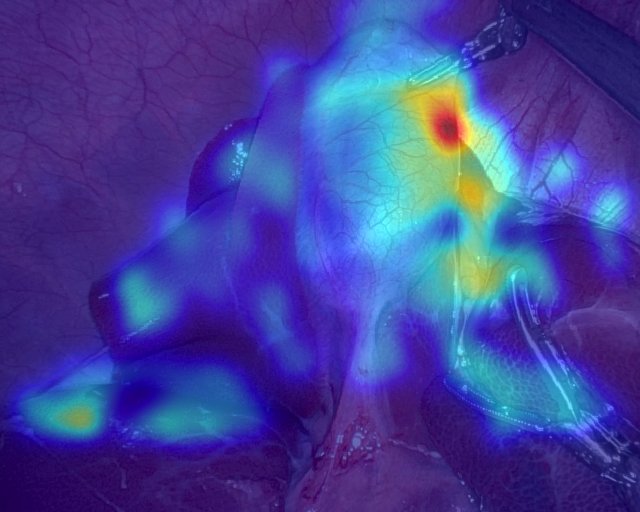} &
\includegraphics[width=0.095\linewidth]{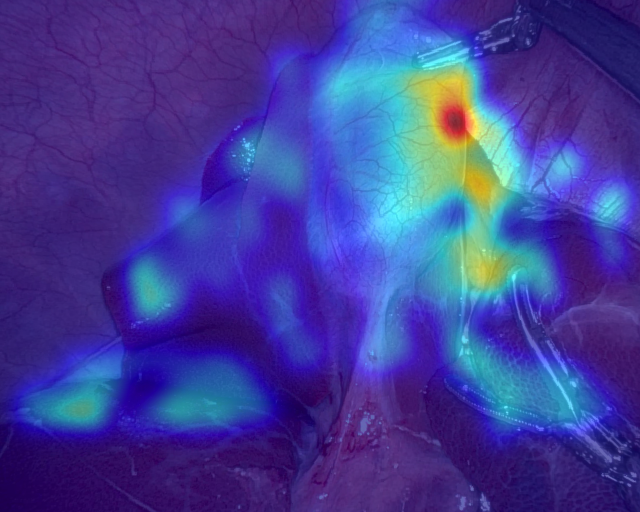}
\end{tabular}

\caption{\textbf{Left to right = increasing time.} Each column is at the same timestep. First row: Raw RGB frames. 
Second row: Baseline PAS. Third row: Compliance score (PACS). 
Red regions have higher affordance values, and blue regions have lower affordance values. The affordance progressively concentrates near the tool-tissue interaction region as retraction begins. 
The PACS results exhibit more stable convergence over time.}

\label{fig:pulling_affordance}
\end{figure*}

\begin{figure}[!htbp]
\centering

\setlength{\tabcolsep}{1pt}        
\renewcommand{\arraystretch}{1}   

\begin{tabular}{cc}
\includegraphics[width=0.48\linewidth]{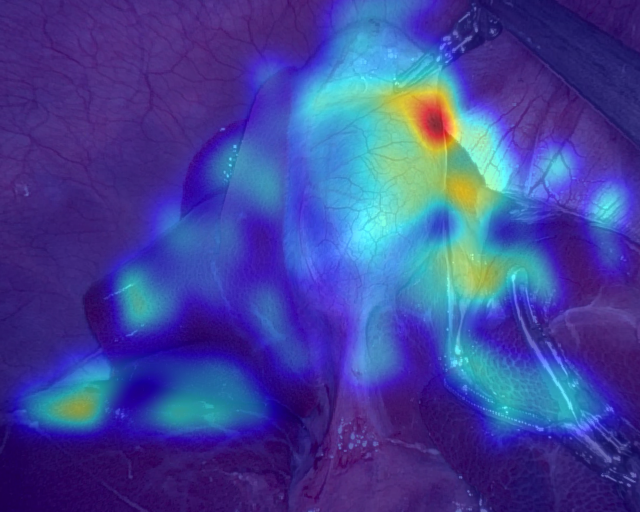} &
\includegraphics[width=0.48\linewidth]{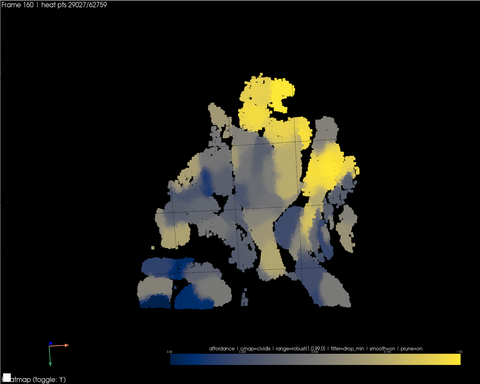}
\\[1pt]

\includegraphics[width=0.48\linewidth]{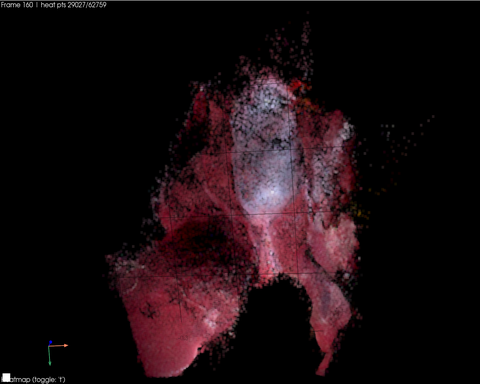} &
\includegraphics[width=0.48\linewidth]{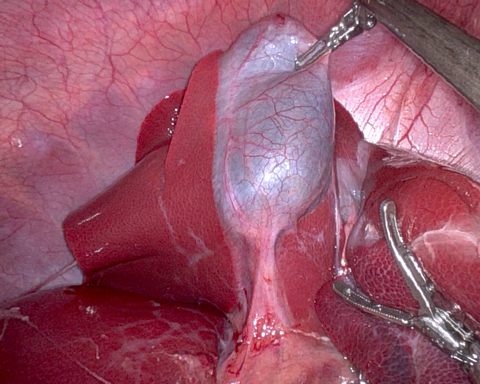}
\end{tabular}

\caption{Qualitative visualization of a sample frame. 
\textbf{Top-left:} Affordance heatmap in 2D. 
\textbf{Top-right:} 3D PACS score distribution heatmap on the tissue point cloud. Yellow: Higher score. Dark blue: Lower score.
\textbf{Lower-left:} reconstructed noisy 3D tissue point cloud. 
\textbf{Lower-right:} original 2D RGB frame.}
\label{fig:frsample}
\end{figure}

\subsection{Affordance Reasoning}

\subsubsection{Empirical Baseline}
\label{sec:exp_pae_baseline}

To assess the benefit of physics-aware compliance modeling, we compare PACS against a purely kinematic co-motion baseline termed \textbf{Positional Agreement Score (PAS)}. This baseline measures statistical alignment between observed tissue motion and the executed tool action direction without incorporating deformation constraints or stiffness structure.

Over a temporal window of length $\tau$, the displacement sequence of a tracked point $\mathbf{x}_i$ is defined as $\mathbf{u}_{t,i} = \mathbf{x}_{t+1,i} - \mathbf{x}_{t,i}$ for $t \in \{t_0, \dots, t_0+\tau-1\}$. The empirical mean displacement is $\bar{\mathbf{u}}_i = \tfrac{1}{\tau}\sum_{k=0}^{\tau-1}\mathbf{u}_{t+k,i}$ and the corresponding covariance is
\[
\mathbf{\Sigma}_{i,t}^{\mathrm{traj}} =
\tfrac{1}{\tau}
\sum_{k=0}^{\tau-1}
(\mathbf{u}_{t+k,i}-\bar{\mathbf{u}}_i)
(\mathbf{u}_{t+k,i}-\bar{\mathbf{u}}_i)^\top .
\]

Let $\mathbf{d}_a \in \mathbb{R}^3$ denote the unit action direction of the tool at time $t$, obtained from the translational component of the incremental tool motion as given in Eq. \ref{eq:tool_induced_displacement}, i.e., $\mathbf{d}_a = \mathbf{v}_t / \|\mathbf{v}_t\|$. \textbf{Positional-Agreement Energy (PAE)} evaluates the directional variance of tissue motion along this action direction:
\begin{equation}
E_{i,t}^{\mathrm{PAE}} =
\mathbf{d}_a^\top
\mathbf{\Sigma}_{i,t}^{\mathrm{traj}}
\mathbf{d}_a .
\end{equation}

PAE therefore identifies regions whose motion statistically co-varies with the tool direction over a short horizon. However, because it relies solely on kinematic correlation and does not encode constraint-induced compliance or stiffness anisotropy, it cannot distinguish mechanically admissible deformation from incidental background motion. We use PAE strictly as a kinematic baseline for quantification.

We evaluated using \textbf{Positional Agreement Score (PAS)}:
\begin{equation}
\mathcal{A}^{PAS}_{i,t}
=
- E^{PAE}_{i,t},
\end{equation}

\subsubsection{Experiments and Results}
\label{sec:exp_res}
Fig.~\ref{fig:pulling_affordance} illustrates the evolution of the pulling action affordance on a gallbladder. Comparing PAS and PACS results, we can see that PACS is a more robust metric for affordance: The affordance heatmaps for PACS evolve stably while the ones for PAS evolve with a high variance. 

From the affordance heatmaps, we can also see the evolution of the interaction compatibility between the surgeon's tool motion and the tissue stiffness distributions (compilant directions):

When the tool is initially static (first figure from the left in Fig.~\ref{fig:pulling_affordance}), the pulling affordance heatmap is broadly distributed and appears almost random. This indicates that many tissue locations satisfy the positional alignment requirements for the pulling action given the current tool-tip pose, and hence multiple regions are potentially manipulable. As the surgeon begins to rotate the gripper (the 3rd and 4th column in the figure) to prepare for pulling the gallbladder, the pulling affordance region shifts toward the area where the rotation occurs. At this stage, while the PAS can capture the affordance to such rotational actions at a few timesteps, the PACS captures such motions more stably and smoothly in the time interval. When the tool is static again (third figure), the PAS regions become more spread out while the PACS regions remains stably converged to the interaction region. Then, as the tool pulls the gallbladder to the right, the affordance region concentrates on the area where the tool interacts with the tissue. This aligns with the intuition that during pulling, the surgeon positions and orients the tool to maximize alignment with the target tissue for saving efforts in pulling.

Figure \ref{fig:frsample} gives qualitative comparions on affordance results in 3D and 2D at the sample frame. In 3D, we can see that the affordance concentrates on the tissue regions where interactions occur, indicating that the current tool motion aligns with tissue compilance, that is, follows the softest direction of the tissue well. We can also see that there are other non-interaction regions where the compilance also aligns with the tool motion, though they are not actually interacted (grasped, in this case) by the tool. 

However, from figure \ref{fig:frsample}, we can also see that the tissue part grasped by the tool lacks affordance values in 3D. Potential reasons include: (1) 3D reconstructions are noisy there because of the tool occlusions, as shown in the lower-left noisy reconstruction results. (2) The furthest-point sampling strategy we use now doesn't sample points closed to the contact region. This motivates contact-aware sampling of representative points.

Fig.~\ref{fig:pulling_affordance} also shows that background tissue regions not in direct contact with the tool exhibit motion regardless of the tool’s activity. This is primarily due to physiological motions such as heartbeats, which introduce random movements and result in noisy, non-actionable affordance regions. Future work could apply temporal smoothing or filtering techniques to mitigate such noise.

\section{Conclusions}
\label{sec:conclusions}
We introduced \methodName, a framework for affordance reasoning in surgical robotics that unifies 3DGS with XPBD-based compliance modeling. By embedding physical constraints into a tracked scene representation, we construct RGPs that capture both geometry and anisotropic deformation behavior. We derive PAS and PACS to assess motion consistency and physical feasibility for tool-tissue interactions. Experiments on surgical video datasets show that \methodName yields more reliable and physically interpretable affordance predictions than visual-only baselines, supporting surgical visual understanding and robotic embodiment.

Future work will (i) improve affordance temporal stability via spatiotemporal regularization, (ii) automate tool pose estimation at scale, (iii) extend to multi-tool interactions and additional procedures, and (iv) couple affordance predictions with closed-loop control for real-time physics-aware assistance for surgical robots.

\bibliographystyle{ieeetr}
\bibliography{references}
\end{document}